\documentclass[runningheads]{llncs}
\usepackage{graphicx}
\usepackage{svg}
\usepackage{amssymb, amsmath}
\usepackage{ifthen}
\usepackage{subcaption}
\usepackage{multirow}
\usepackage[breaklinks=true]{hyperref}
\usepackage{color}

\hypersetup{
    colorlinks=true,
    citecolor=blue,
    linkcolor=blue,
    filecolor=blue,      
    urlcolor=blue
    }

\usepackage{tikz}
\definecolor{lightgreen}{HTML}{D5E8D4}
\definecolor{lightred}{HTML}{F8CECC}

\newcommand*{\tikzbullet}[1]{%
  \setbox0=\hbox{\strut}%
  \begin{tikzpicture}
    \useasboundingbox (-.25em,0) rectangle (.25em,\ht0);
    \filldraw[draw=#1,fill=#1] (0,0.5\ht0) circle[radius=.25em];
  \end{tikzpicture}%
}
\newcommand*{\tikzsquare}[1]{%
  \setbox0=\hbox{\strut}%
  \begin{tikzpicture}
    \useasboundingbox (-.25em,0) rectangle (.25em,\ht0);
    \draw[draw=#1] (-.25em,0) rectangle (.25em,\ht0);
  \end{tikzpicture}%
}

\newcommand*{\roundtikzsquare}[1]{%
  \setbox0=\hbox{\strut}%
  \begin{tikzpicture}
    \useasboundingbox (-.25em,0) rectangle (.25em,\ht0);
    \draw[draw=#1, fill=#1, rounded corners=1pt] (-.25em,0) rectangle (.25em,\ht0);
  \end{tikzpicture}%
}

\begin{document}
\title{SAM Carries the Burden: A Semi-Supervised Approach Refining Pseudo Labels for Medical Segmentation}
\titlerunning{SAM Carries the Burden}

\newboolean{anonymized}
\setboolean{anonymized}{False}

\ifthenelse{\boolean{anonymized}}{%
    \author{
        Anonymized Authors
    }
    \authorrunning{Anonymized Author et al.}
    %
    \institute{Anonymized Institute\\
        \email{author@institute.com}\\
    }
}{%
    \author{Ron Keuth\inst{1}\orcidID{0009-0003-4289-2836} \and
    Lasse Hansen\inst{2}\orcidID{0000-0003-3963-7052} \and
    Maren Balks\inst{3} \and
    Ronja Jäger\inst{3} \and
    Anne-Nele Schröder\inst{3} \and
    Ludger Tüshaus\inst{3} \and
    Mattias Heinrich\inst{1}\orcidID{0000-0002-7489-1972} 
    }
    \authorrunning{Keuth et al.}
    %
    \institute{Institute of Medical Informatics, Universität zu Lübeck, Lübeck, Germany
        \email{r.keuth@uni-luebeck.de} \and
        EchoScout GmbH, Lübeck, Germany \and
        Paediatric Surgery, University Hospital Schleswig-Holstein, Lübeck, Germany}
}
\maketitle              
\begin{abstract}
Semantic segmentation is a crucial task in medical imaging. Although supervised learning techniques have proven to be effective in performing this task, they heavily depend on large amounts of annotated training data.
The recently introduced Segment Anything Model (SAM) enables prompt-based segmentation and offers zero-shot generalization to unfamiliar objects. In our work, we leverage SAM's abstract object understanding for medical image segmentation to provide pseudo labels for semi-supervised learning, thereby mitigating the need for extensive annotated training data.

Our approach refines initial segmentations that are derived from a limited amount of annotated data (comprising up to 43 cases) by extracting bounding boxes and seed points as prompts forwarded to SAM. Thus, it enables the generation of dense segmentation masks as pseudo labels for unlabelled data.
The results show that training with our pseudo labels yields an improvement in Dice score from $74.29\,\%$ to $84.17\,\%$ and from $66.63\,\%$ to $74.87\,\%$ for the segmentation of bones of the paediatric wrist and teeth in dental radiographs, respectively.
As a result, our method outperforms intensity-based post-processing methods, state-of-the-art supervised learning for segmentation (nnU-Net), and the semi-supervised mean teacher approach.
Our Code is available on \href{https://github.com/multimodallearning/SamCarriesTheBurden}{GitHub}.

\keywords{Segmentation  \and Segment Anything Model \and Semi-Supervised Learning}
\end{abstract}
\section{Introduction}
Semantic segmentation plays a pivotal role in medical imaging, as it is fundamental for the identification and further analysis of objects of interest in images.
While supervised learning techniques have demonstrated remarkable success in automating image segmentation tasks \cite{isensee_nnu-net_2021}, their efficacy hinges upon the availability of large datasets with high-quality annotations.
Manually segmenting images is time-consuming, as it requires precise labelling of the individual anatomical boundaries at pixel level.
The complexity is particularly high in the medical field and requires specialised knowledge, which drives up the associated costs. Specifically, in our work on paediatric wrist bone and teeth segmentation, challenges arise from the overlapping nature of carpal bones and teeth in 2D radiographs and the potential absence of classes. For the wrist setting, the latter is caused by the delayed ossification of individual carpal bones during the first years of childhood. For the teeth segmentation, it results from tooth loss at advanced age. 
Additionally, wrist bone segmentation plays an important role for determining the bone age \cite{Adeshina_bone_seg,faisal2021x}.\par
Our work is motivated by two key observations:
First, we observe that the accuracy of predicted segmentation masks in supervised methods is correlated with the amount of annotated training data.
However, even a few training examples can yield a model capable of providing robust but partially imprecise segmentations, capturing the domain knowledge like bones constellation of the hand skeleton.
Second, the recent introduction of foundation models has enabled prompt-based segmentation, offering robust zero-shot generalization to unfamiliar objects.
These models provide an abstract understanding of objects, leading to accurate segmentation maps.\\
\textbf{Contribution:}
Building on these insights, we propose a method that leverages initial imprecise segmentation masks to provide domain knowledge within prompt-based segmentation foundation models.
Our pipeline enables the Segment Anything Model (SAM) for pseudo label refinement in a semi-supervised setting. With our automatic generation of prompting, SAM refines initial segmentation masks derived from a limited amount of labelled data and thus provides pseudo labels, enabling unlabelled data for training.
We empirically demonstrate on two medical segmentation tasks that training with refined pseudo labels leads to superior segmentation performance compared state-of-the-art supervised (nnU-Net) and semi-supervised (Mean Teacher) methods.

\subsection{Related Work}
Motivated by the costs, a wide variety of methods has emerged, aiming to facilitate the annotation process for semantic segmentation in the medical domain by either minimizing the amount of labelled data needed for training (semi-supervised learning) or developing tools aiding human experts to speed up the annotation process (semi-automatic annotation).\\
\textbf{Semi-supervised learning:} It is an active field of research \cite{Jiao2022LearningWL}, which leverages similarities between labelled and unlabelled data to improve generalization performance.
This involves generating pseudo labels \cite{lee_pseudo_label} supported by domain prior knowledge \cite{Zheng_prior}, or utilizing unsupervised regularization methods, e.g. consistency loss functions for unlabelled data. \cite{laine2017temporal,tarvainen2017mean}\\
\textbf{Semi-automatic annotation:} Prior knowledge to distinguish an object from its environment can be used to support the human expert during the annotation process. The techniques used can be based on image intensities \cite{Grady2006RandomWF} or graphs \cite{Rother2004GrabCutIF}, but could also be learning-based \cite{diaz2022monai,Kirillov2023SegmentA,wang2018interactive}.\\
\textbf{SAM:} With the introduction of the Segment Anything Model (SAM) \cite{Kirillov2023SegmentA}, learning-based annotation has gained attention \cite{Zhang2024SegmentAM}.
SAM enables prompt-based interactive segmentation and offers zero-shot generalization to unfamiliar objects, including the medical image domain with a variety of modalities \cite{Huang2024}.
While recent efforts have primarily focused on adapting and fine-tuning SAM to the medical domain \cite{Cheng2023SAMMed2D,ma2024segment}, other work is exploiting SAM for a synergy with semi-supervised learning. For instance, \cite{Chen2023SegmentAM} utilize class activation maps (CAM) to provide the rough object location as prompts for SAM to predict its accurate segmentation mask.
\cite{Huang2023PushTB} introduces a method where SAM generates pseudo labels via self-prompting, followed by selection and refinement using a self-label correction module.
Other approaches leverage SAM to identify correct pseudo labels generated by a segmentation model trained on limited annotated data, achieved through agreement between predictions \cite{Li2023SegmentAM,Zhang2023SamDSKCS} or by integrating SAM into existing semi-supervised frameworks \cite{Zhang2023SemiSAMES}.\par
All methods show promise but face limitations in both our settings: 
CAM requires positive and negative examples of each class for proper distinction, which would require a patch-based training where most classes are present in every image.
\cite{Huang2023PushTB} manually places the same seed points as prompts in every image, which requires common locations of all anatomies across the dataset. However, this is not fulfilled for our datasets as they lack pre-alignment.
In our settings, selecting reliable pseudo labels based on agreement \cite{Li2023SegmentAM} rather than refining them only slightly increases the number of high-quality labels available for training.
While including SAM on-the-fly in the Mean Teacher \cite{Zhang2023SemiSAMES} bears the risk of wrong guidance.

\section{Methods}
In a semi-supervised segmentation problem, we have a dataset $\mathcal{D}$ that holds a set of images $\mathcal{I}$, where only a small subset of images $\mathcal{X}$ has corresponding segmentation masks $(\mathcal{X}, \mathcal{S})$.
Further, we define a disjunct training $(\mathcal{X}_\text{train}, \mathcal{S}_\text{train})$, validation $(\mathcal{X}_\text{val}, \mathcal{S}_\text{val})$ and test subset $(\mathcal{X}_\text{test}, \mathcal{S}_\text{test})$, where the amount of unlabelled data in the subset $\mathcal{Y}=\mathcal{D}\setminus \mathcal{X}$ is much greater (${|\mathcal{Y}|\ll|\mathcal{X}|}$).
Given $\mathcal{D}$, our aim is to train a model that generates accurate segmentation maps for a given image with the supervision of $(\mathcal{X}_\text{train}, \mathcal{S}_\text{train})$ but also exploits the unlabelled data in $\mathcal{Y}$ to further improve its generalization ability.

\begin{figure}[h]
    \centering
    \includesvg[width=\textwidth]{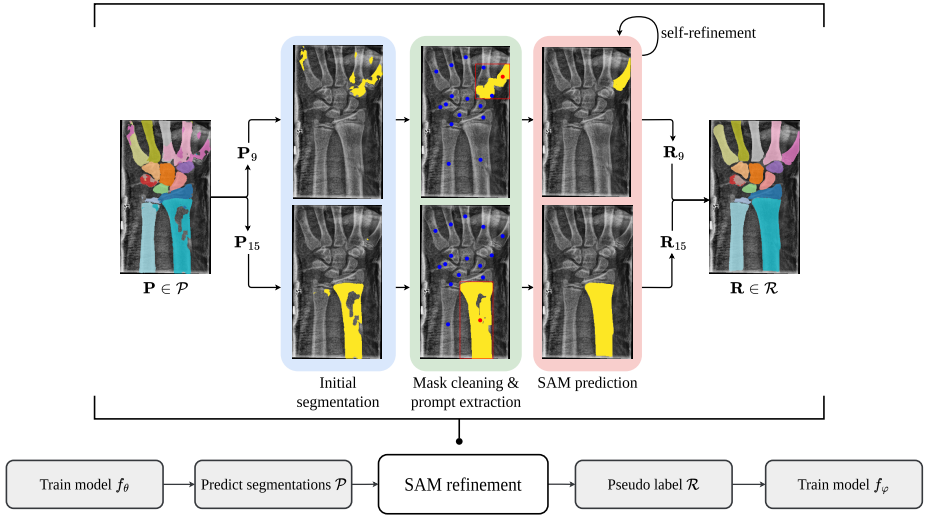}
    \caption{Proposed pipeline: We use SAM to refine predictions $\mathcal{P}$ of the segmentation model $f_\theta$ trained on few labelled data. After mask cleaning, we extract a bounding box (\protect\tikzsquare{red}), one positive (\protect\tikzbullet{red}) and multiple negative (\protect\tikzbullet{blue}) seed points as sparse prompts for SAM, obtaining refined pseudo labels $\mathcal{R}$. Finally, we train a new segmentation model $f_\varphi$ with $\mathbf{R}$ on unlabelled data. See Sec. \ref{sec:sam_pseudo_label} for details.}
    \label{fig:method_overview}
\end{figure}

\subsection{Pseudo Labelling with SAM} \label{sec:sam_pseudo_label}
Due to the lack of labels in $\mathcal{Y}$, we ensure the quality of the predicted segmentation masks by leveraging SAM's abstract object understanding via automatically prompt extraction.
Fig. \ref{fig:method_overview} presents our proposed training routine consisting of five steps:
1) Train segmentation model $f_\theta$ on few labelled data ($\mathcal{X}_\text{train}, \mathcal{S}_\text{train})$.
2) Predict $\mathbf{P} = f_\theta(\mathbf{I}\in\mathcal{Y})$, creating a paired set of image and predicted segmentation $(\mathcal{Y},\mathcal{P})$.
3) Preprocess $\mathcal{P}$ (mask cleaning) and extract prompts for SAM.
4) Employ SAM to generate a refined version $\mathbf{R} = \texttt{SAM}(\mathbf{P})$ for $\mathbf{P}$ and obtain a pair of images and pseudo labels $(\mathcal{Y}, \mathcal{R})$.
5) Train segmentation model $f_\varphi$ on $(\mathcal{Y}, \mathcal{R})$.

\subsubsection{Mask Cleaning and Prompt Extraction.} \label{sec:prompt_extract_preprocessing}
SAM takes axis aligned bounding boxes and seed points as sparse prompts. The latter are differentiated between positive and negative ones describing the object or background.
We generate those prompts from the model's predicted segmentation $\mathbf{P}\in\mathcal{P}$ by iterating over the binary mask $\mathbf{P}_i$ of each class $\omega_i\in\Omega$ and extracting its bounding box as well as its center of mass as a positive seed point (Fig \ref{fig:method_overview}, \tikzbullet{red}), while the seed points of all other classes $\{\omega_j\in\Omega|\omega_j\neq\omega_i\}$ act as negative ones (Fig. \ref{fig:method_overview}, \tikzbullet{blue}).
Due to the possible residual errors in $\mathbf{P}_i$, applying preprocessing steps for mask cleaning before prompt extraction is crucial. First, we utilize a connected component analysis for each class $\omega_i$ eliminating all but the component $C^*_i\in\mathcal{C}_i$ with the highest predicted likelihood normalized by its area:
\begin{equation}
    C^*_i=\underset{C_i\in\mathcal{C}_i}{\arg\max}\frac{\sum_{\hat{p}\in C_i}\hat{p}}{|C_i|},
\end{equation}
where $C_i$ holds all pixels of the component with their predicted likelihood $\hat{p}$ of segmentation class $\omega_i$. 
After eliminating all but one component, we apply morphological operations to all $\{C^*_i\}_{i=1}^{|\Omega|}$ by choosing from erosion or dilation using different structure elements and radii to ensure robustness of the prompt extraction.

\subsubsection{Self-refinement}\label{sec:self_refinement}
As the final refinement step, we give SAM its previous mask prediction as a dense prompt. When combined with sparse prompting (e.g. seed points), it improves the homogeneity of generated segmentations (Fig. \ref{fig:method_overview}, \roundtikzsquare{lightred}).

\section{Experiments}
\subsection{Experimental Setup}
\subsubsection{Datasets.}\label{sec:datasets}
We evaluate our method on two medical segmentation tasks:
The first is segmenting bones of paediatric wrists using the publicly available\\ \mbox{GRaZPEDWRI-DX} dataset \cite{nagy_pediatric_2022} (project No. EK 31-108 ex 18/19; IRB-approved by IRB00002556; CC BY 4.0).
Due to the high overlap of the bones, we only include anterior view and normalise the orientation to left laterality, obtaining dataset $\mathcal{D}_w$ with 10k images of 5900 patients. From it, we picked 62 images, that represent the distributions of age, gender, pathologies, and disturbances (foreign objects or plaster) of $\mathcal{D}_w$.
Our radiologists manually annotated the segmentation masks for $17=|\Omega|$ bones.
We split the dataset over the number of patients into 43 training, 10 validation, and 9 test cases.
For our unlabelled data subset $\mathcal{Y}$, we sample 500 images randomly from $\mathcal{D}_w\setminus \mathcal{X}$.\par
As a second dataset, we employ the publicly available teeth segmentation dataset on dental radiographs $\mathcal{D}_d$ \cite{dental_dataset} (CC0 1.0) including 598 images with segmentation masks of up to 32 teeth annotated by 12 human experts.
We use a custom split of 450 training, 75 validation and 73 test images.
To fit the semi-supervised setting of our first dataset, we consider the first 45 training data as labelled $\mathcal{X}_\text{train}$ and the remaining 405 as unlabelled $\mathcal{Y}$.

\subsubsection{Metric.}
We evaluate our methods using the Dice Similarity Coefficient (DSC), calculating it for each radiograph and average it across the test split.

\subsubsection{Ablation Study.}
To demonstrate the effectiveness of our method, we sample more subsets of the paediatric wrist dataset $\mathcal{X}_\text{train}\in\mathcal{D}_w$ with successively fewer training data $\{\mathcal{X}_\text{train}^n\}_{n=1}^{|\mathcal{X}_\text{train}|}$, where $n$ increases with a step size of 5 and describes the amount of data in $\mathcal{X}_\text{train}^n$. For $\mathcal{X}_\text{train}^1$ a single training image covering all classes is manually picked. The data for all further subsets are added by randomly sampling of the remaining unused training data.

\subsubsection{Implementation Details.}\label{sec:impl_details}
We use the U-Net \cite{ronneberger_2015} (depth 4, 64 channels, 31M parameters) as the segmentation model $f$ and train it with Adam optimizer, minimising the weighted binary cross entropy (BCE) loss with an initial learning rate of $1e^{-3}$ reduced by factor 100 with cosine annealing over 1050 iterations with batch size 16. The training requires 11GB VRAM and up to 2 hours on an RTX 2080 Ti.
For data augmentation, we apply random affine transformation with parameters sampled from a Gaussian distribution $\mathcal{N}(0, 0.03)$.
For SAM and MedSAM, we use the off-the-shelf models without any further adaption or fine-tuning.
For SAM refinement process, we optimize the choice of prompts (including the use of self-refinement) on the validation split as well as for the morphological operator its structure element and its radius in pixels $\mathcal{U}(1,8)$ using the Tree-Structured Parzen estimator (TPE) approach (Optuna \cite{optuna} implementation).
Our code, including 63 manual segmentations by our radiologists of the paediatric wrist dataset, are available on \href{https://github.com/multimodallearning/SamCarriesTheBurden}{GitHub}.

\subsubsection{Comparison Methods.}
\textbf{MedSAM} \cite{ma2024segment} provides an adaption of SAM to the medical domain.
This model is limited to box-prompts only and lacks the ability of self-refining.
For an intensity-based refinement approach, we include the \textbf{Random Walk} algorithm \cite{Grady2006RandomWF} and optimize its hyperparameters over $(\mathcal{X}_\text{val}, \mathcal{S}_\text{val})$ in the same way as for SAM (Sec. \ref{sec:impl_details}).
As a baseline for supervised learning, we extend the \textbf{nnU-Net} framework \cite{isensee_nnu-net_2021} with a binary cross entropy (BCE), enabling the training with overlapping labels, and train one model for 100 epochs without cross validation leaving all other framework parameters untouched.
To investigate SAM's general object understanding, we use its ViT image encoder with an \textbf{LR-ASPP head} \cite{howard_searching_2019}.
The LR-ASPP considers ViT's image embedding as "intermediate" features, while we employ two residual blocks \cite{he_resnet_2016} to extract the high-level features.
During training, we leave the ViT fixed, only optimizing the LR-ASPP head and the two residual blocks.\par
For a comparison to semi-supervised methods, we chose the \textbf{Mean Teacher} approach \cite{tarvainen2017mean}.
We use the trained U-Net $f_\theta$ as starting point and apply the same training routine as in Sec. \ref{sec:impl_details}, choosing the EMA-coefficient for the teacher update of $\alpha_\text{MT}=0.996$.
For a comparison also utilizing SAM, we adapt \cite{Li2023SegmentAM} to our multilabel setting and expand our Mean Teacher with its SAM-based selection of reliable pseudo labels via agreement. Setting the Dice threshold $\alpha_\text{SAM}$ to $85\,\%$ and $65\,\%$ for our wrist and teeth task, respectively.

\subsection{Results}
\begin{table}[t]
    \caption{Quantitative results on the test splits of our two datasets, reported as Dice scores ($\mu_{\pm\sigma}$) in Percent. The symbols $f_\theta, \mathcal{P}, \mathcal{R}$ reference to Fig. \ref{fig:method_overview}.}
    \centering
    \setlength{\tabcolsep}{.3cm}
    \bgroup
    \def\arraystretch{1.}
    \begin{tabular}{llcc}
        \hline
        & & \multicolumn{2}{c}{Dataset}\\
        Supervision & Method & Wrist & Teeth\\\hline
        \multirow{3}{*}{Fully} & nnU-Net\cite{isensee_nnu-net_2021} BCE & $77.6_{\pm14.6}$ & $67.5_{\pm14.5} $ \\
        & SAM ViT + LR-ASPP\cite{howard_searching_2019} & $71.8_{\pm6.2}$ & $45.2_{\pm3.8}$ \\
        & U-Net\cite{ronneberger_2015} ($f_\theta$)& $74.3_{\pm14.4}$ & $66.6_{\pm7.7}$ \\\hline
        \multirow{3}{*}{Refinement}&U-Net + SAM\cite{Kirillov2023SegmentA} & $80.6_{\pm15.9}$ & $\mathbf{75.8_{\pm12.2}}$ \\
        & U-Net + MedSAM\cite{ma2024segment} & $70.5_{\pm17.3}$ & $69.8_{\pm8.3}$ \\
        & U-Net + Random Walk\cite{Grady2006RandomWF} & $74.7_{\pm15.0}$ & $67.1_{\pm8.0}$ \\\hline
        \multirow{2}{*}{Semi}& Mean Teacher\cite{tarvainen2017mean} (MT) & $79.3_{\pm4.7}$ & $62.2_{\pm8.1}$ \\
        &MT + SAM selection\cite{Li2023SegmentAM} & $68.3_{\pm23.3}$ & $63.0_{\pm7.9}$\\\hline
        \multirow{2}{*}{Pseudo} & ours w/o SAM ($\mathcal{P}$)& $74.7_{\pm8.7}$ & $58.9_{\pm5.6}$ \\
        & \textbf{ours w/ SAM} ($\mathcal{R}$) & $\mathbf{84.2_{\pm8.0}}$ & $74.9_{\pm9.4}$ \\\hline
    \end{tabular}
    \egroup
    \label{tab:quantitative results}
\end{table}

With the hyperparameter optimisation on our validation split, we determine SAM performs best with an initial bounding box prompting followed by self-refinement utilizing positive and negative seed points as additional prompts. For the preprocessing of the initial provided segmentation mask $\mathbf{P}$, dilation with an isotropic squared structure element of size 8 (image sizes $384\times224$ and $224\times384$) showed the highest robustness.\par
Quantitative results for the test splits of both datasets are shown in Tab. \ref{tab:quantitative results}.
When trained with limited annotated data, the nnU-Net outperforms our U-Net and the SAM ViT + LR-ASPP, with the latter performing the worst.
For the refinements of our U-Net's prediction, our method using SAM surpasses the Random Walk and MedSAM yielding an improvement of $6.27\,\%$ DSC for the wrist and $9.18\,\%$ for the dental dataset.
While the Mean Teacher performs second-best in the wrist setting, it fails on the dental dataset.
Extending the Mean Teacher paradigm with SAM for selecting reliable pseudo labels \cite{Li2023SegmentAM} showed no benefits in our settings.
Further training with pseudo labels on unlabelled data $\mathcal{Y}$, the non-refined variant ($\mathcal{P}$) showed no benefits over initial supervised training.
However, refining pseudo labels with SAM ($\mathcal{R}$) improves performance by $9.88\,\%$ DSC for the wrist setting and $8.24\,\%$ for the dental dataset.
This achieves the best performance with $84.17\,\%$ in the wrist setting.
For the dental setting, the performance of $74.87\,\%$ matches supervised training using all 450 annotated data, which reaches $74.19 \pm 8.59\,\%$ (not shown in Tab. \ref{tab:quantitative results}).

\begin{figure}
    \centering    
    \includegraphics[width=\textwidth]{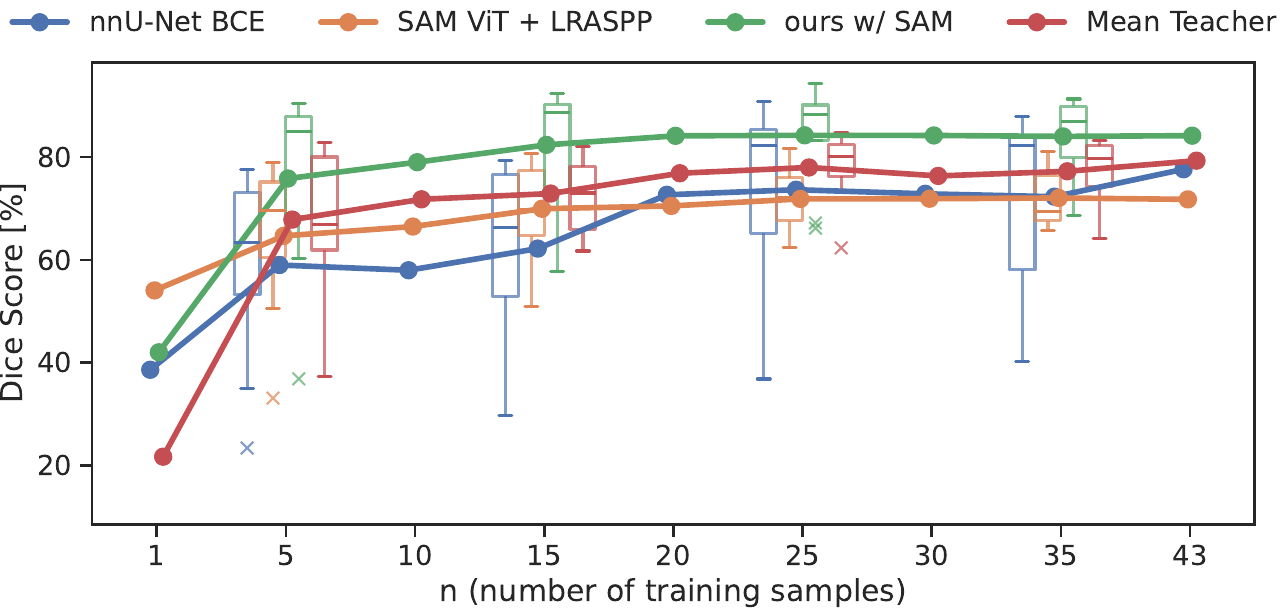}
    \caption{Dice scores reached by a selection of methods trained with an increasing number of labelled data $\mathcal{X}_\text{train}^n$ (see Sec. \ref{sec:datasets} for details). Line plots show the Dice means. We omit the box plots for some $n$ to increase readability.}
    \label{fig:dsc_num_train}
\end{figure}

In Fig. \ref{fig:dsc_num_train}, we show the quantitative results of our ablation study on the wrist dataset.
Here, our U-Net, trained with pseudo labels refined with SAM ($\mathcal{R}$), outperforms the comparison methods regardless of initial training data available.

\begin{figure}
    \centering
    \begin{subfigure}[t]{.23\textwidth}
        \includegraphics[width=\textwidth]{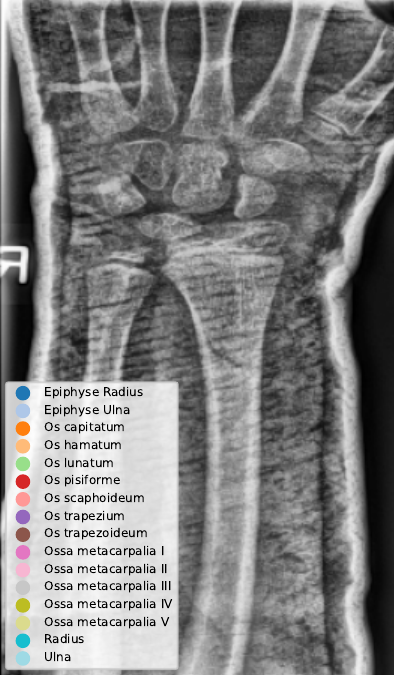}
    \end{subfigure}\hfill
    \begin{subfigure}[t]{.23\textwidth}
        \includegraphics[width=\textwidth]{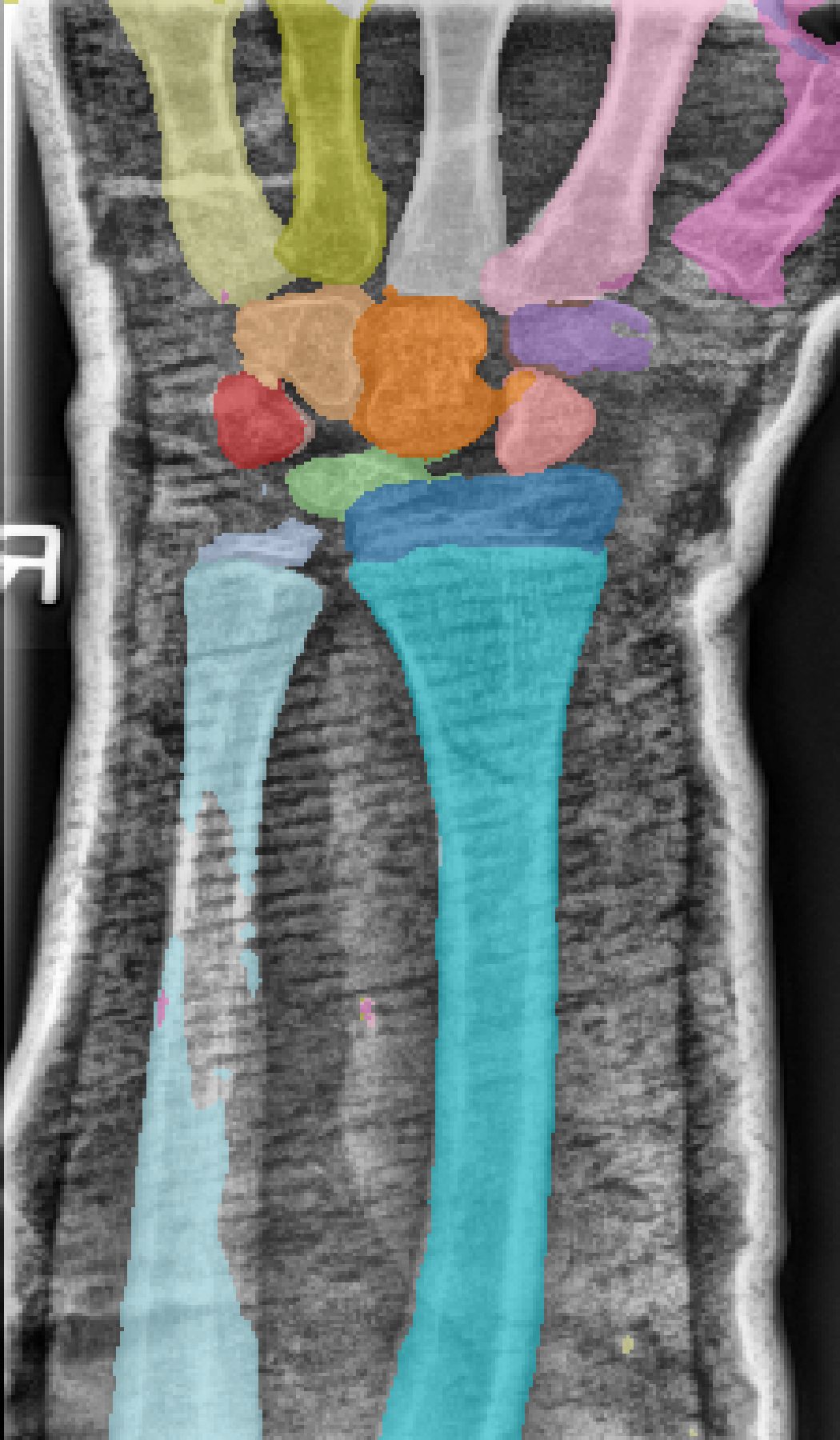}
        \caption{U-Net: $78\pm10\,\%$}
    \end{subfigure}\hfill
    \begin{subfigure}[t]{.23\textwidth}
        \includegraphics[width=\textwidth]{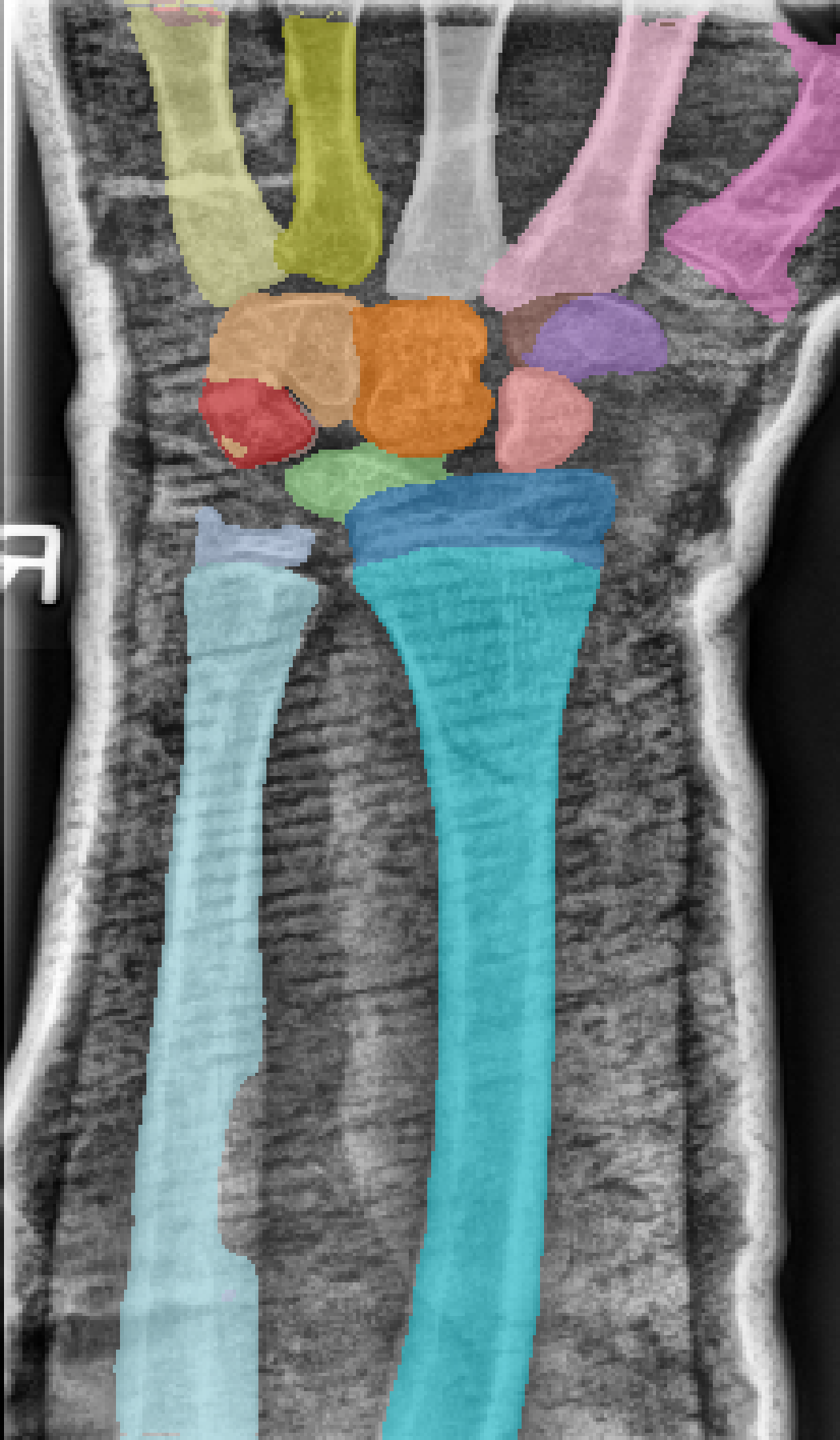}
        \caption{MT: $82\pm9\,\%$}
    \end{subfigure}\hfill
    \begin{subfigure}[t]{.23\textwidth}
        \includegraphics[width=\textwidth]{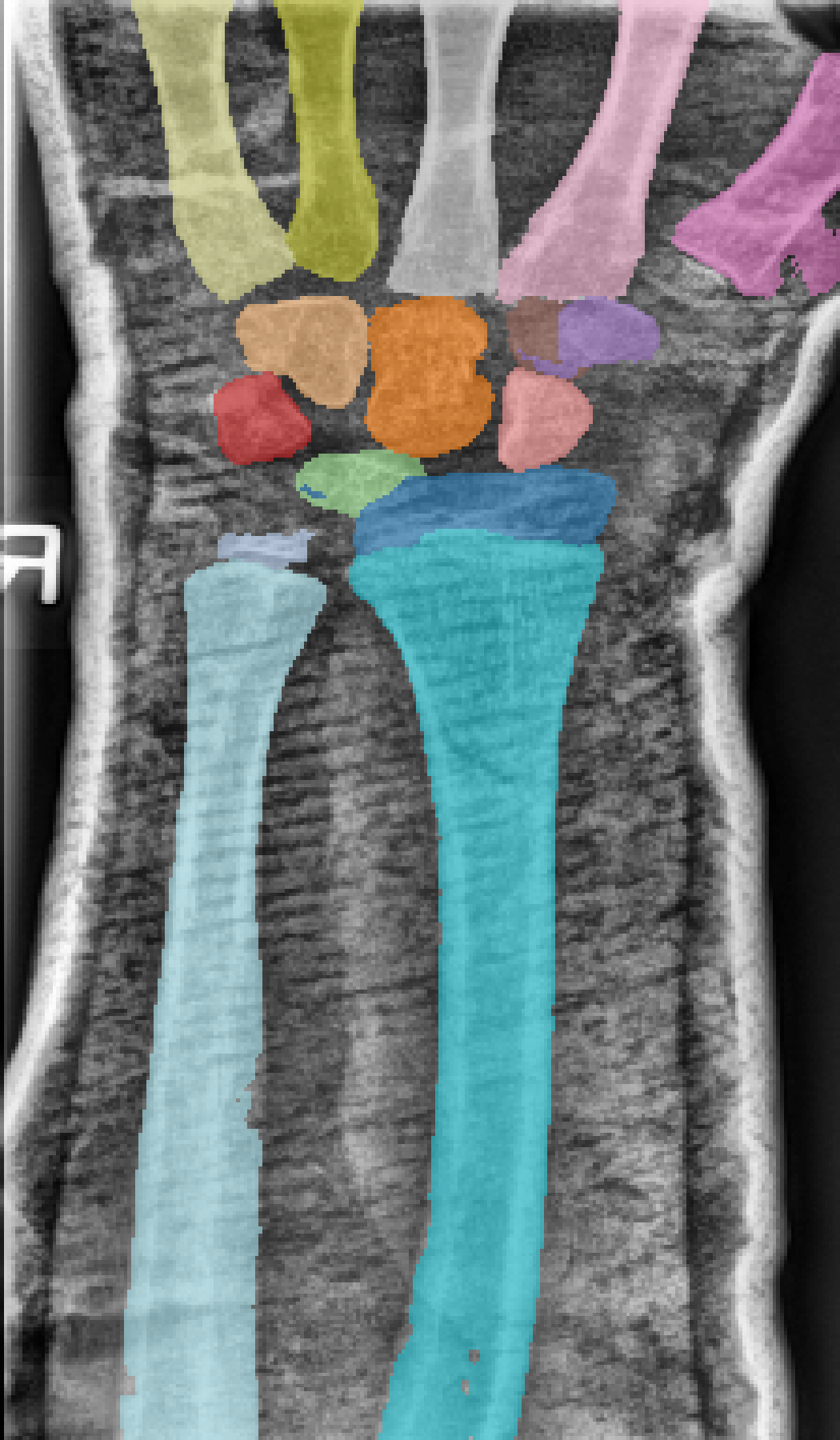}
        \caption{ours: $86\pm13\,\%$}
    \end{subfigure}
    \caption{The median result on wrist test split, including DSC ($\mu\pm\sigma$) for a selection of methods. a) U-Net $f_\theta$, b) Mean Teacher (MT) and c) U-Net $f_\varphi$ trained with SAM refined pseudo labels $\mathcal{R}$ on unlabelled data $\mathcal{Y}$ (ours).}
    \label{fig:qualitative_results}
\end{figure}

The qualitative results in Fig. \ref{fig:qualitative_results} show a challenging test case of the wrist dataset with low contrast due to cast artefacts. 
While the training with unrefined pseudo labels and Mean Teacher suffers from overestimation and heterogeneous segmentation mask, the U-Net trained with SAM refined pseudo labels $\mathcal{R}$ provides accurate segmentation masks.
For further qualitative results, including those from the dental dataset, we refer to the supplemental material.

\section{Discussion and Conclusion}
Our results (Tab. \ref{tab:quantitative results}) demonstrate that the combination of domain knowledge provided by initial segmentations and SAM's abstract object understanding yields high-quality pseudo labels comparable to ground truth ones.
This was evident in the dental dataset, where our method (trained on 45 labelled and 405 unlabelled images) achieved a DSC of $74.9\pm9.4$, compared to $74.2 \pm 8.6\,\%$ trained with 450 labelled images (not shown in Tab. \ref{tab:quantitative results}).
Due to the successful utilization of unlabelled data, our method reaches top performance with relative few initial labelled data (20), as shown in our ablation study (Fig. \ref{fig:dsc_num_train}).
To investigate SAM's zero-shot capabilities, we employed an LR-ASPP onto its frozen ViT, but its poor performance suggests SAM's image embedding alone may not provide its abstract object understanding. 
We could not find any benefit in replacing SAM with its medical-specific version MedSAM, likely due to MedSAM's lack of self-refinement (Sec. \ref{sec:self_refinement}), an important feature of SAM, we adopt in our method improving DSC by $2.21\,\%$ for the wrist dataset.
As a \textbf{limitation}, our method is restricted to semantic segmentation with one instance per class per image due to our mask cleaning.
Also, the initial segmentation for prompt extraction must partially cover the object of interest.
\subsection{Conclusion} In this work, we presented a pipeline utilizing SAM's abstract object understanding refining initial inaccurate segmentation masks via automatic prompt extraction, thus substantially accelerating the annotation of new datasets.
Our results showed, that training with those refined segmentation masks as pseudo labels in a semi-supervised setting can match the performance of using ground truth annotations.
Our approach of refining pseudo labels motivates its application to other fields like domain adaptation, where semi-supervised methods such as the Mean Teacher have already shown potential \cite{Bigalke2023}.

\subsubsection{\ackname}
This research has been funded by the state of Schleswig-Holstein, Grant Number $220\,23\,005$.


%
%
%
\bibliographystyle{splncs04}
\bibliography{sam_carries_the_burden_bibliography}

\newpage

\begin{center}
    \huge \textbf{Supplementary Material: SAM Carries the Burden}
\end{center}

\begin{figure}[h]
    \centering
    \begin{subfigure}[t]{.23\textwidth}
        \includegraphics[width=\textwidth]{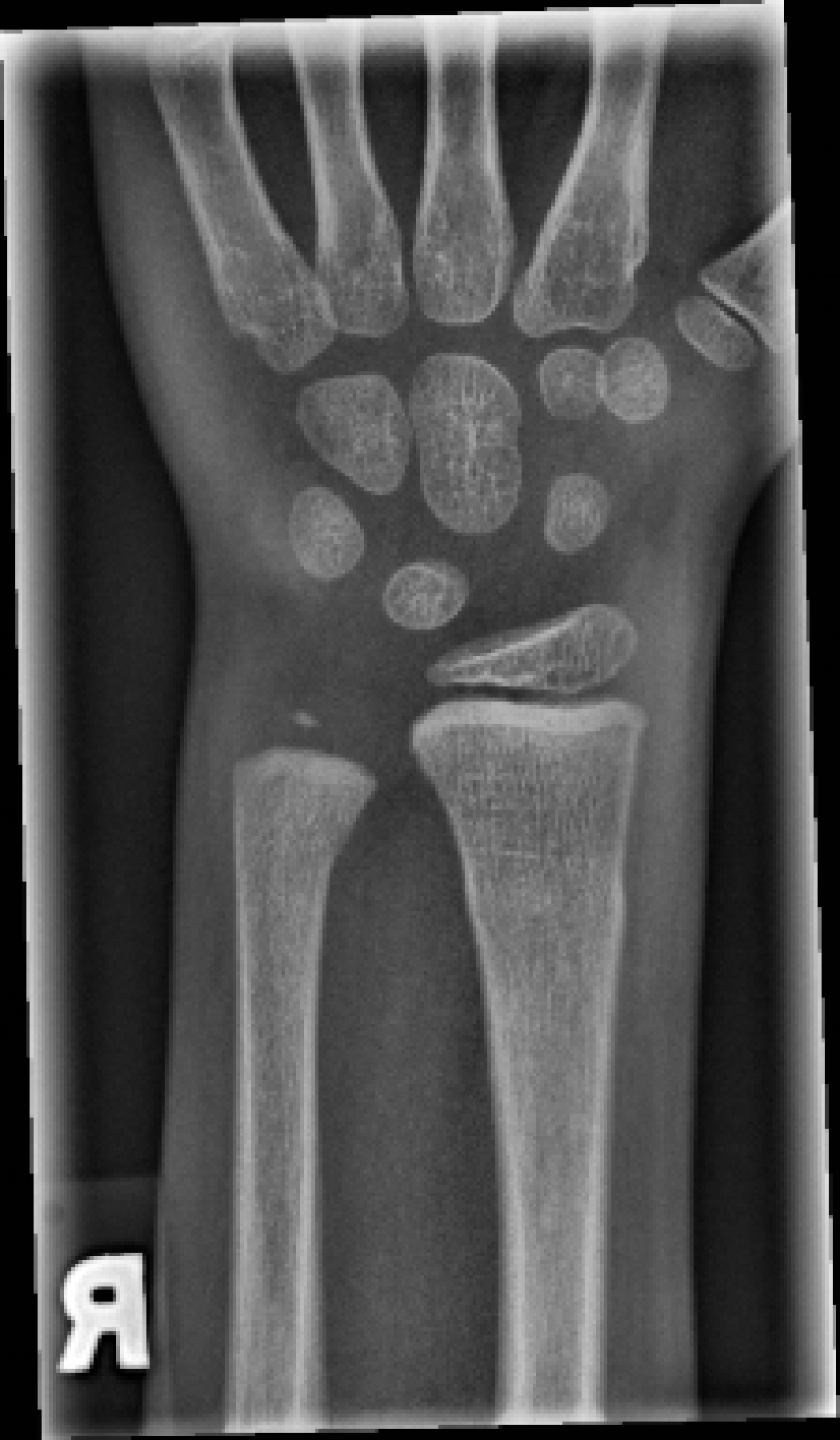}
        \caption{Best test case}
    \end{subfigure}\hfill
    \begin{subfigure}[t]{.23\textwidth}
        \includegraphics[width=\textwidth]{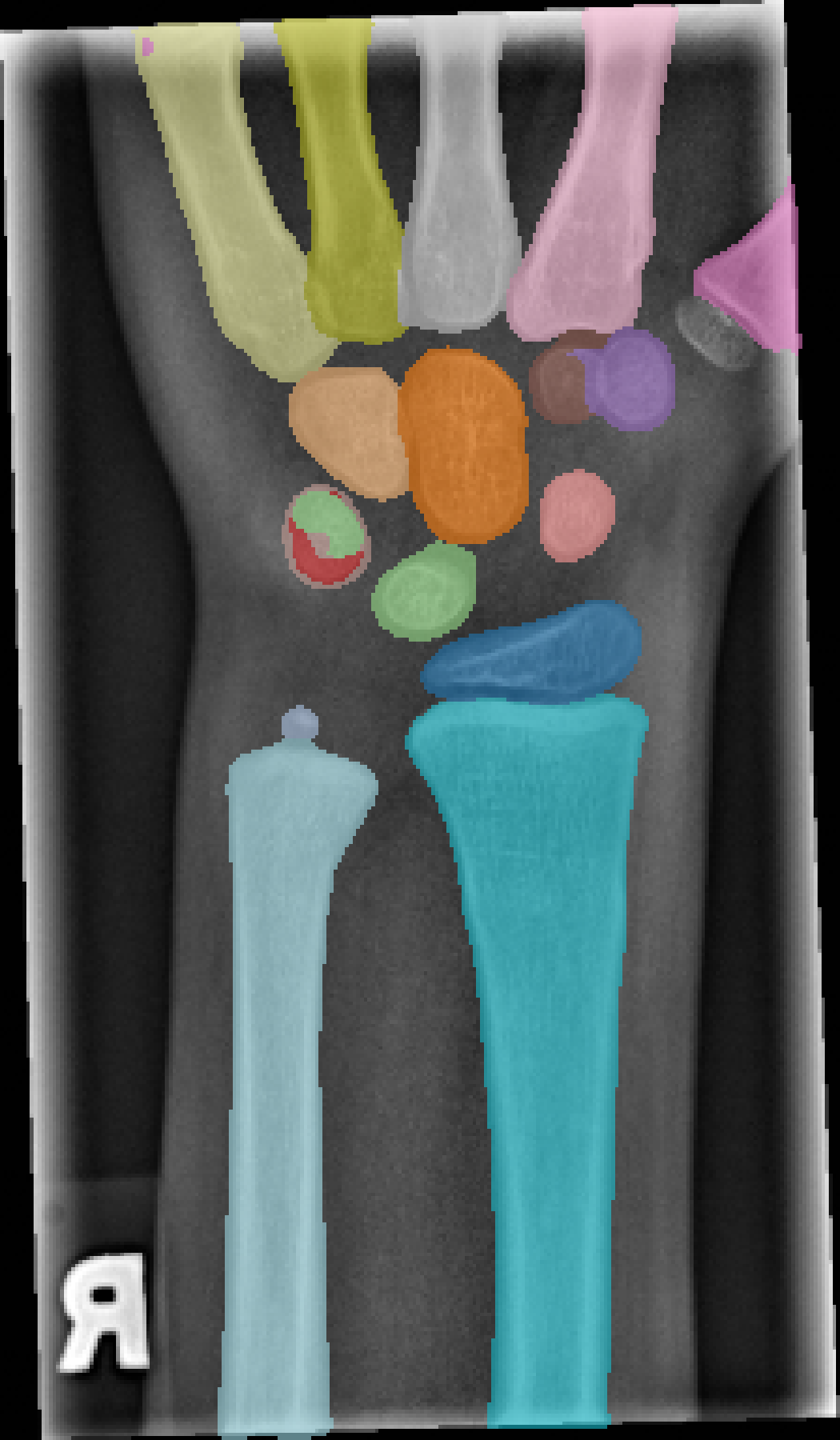}
        \caption{UNet: $79\pm13\,\%$}
    \end{subfigure}\hfill
    \begin{subfigure}[t]{.23\textwidth}
        \includegraphics[width=\textwidth]{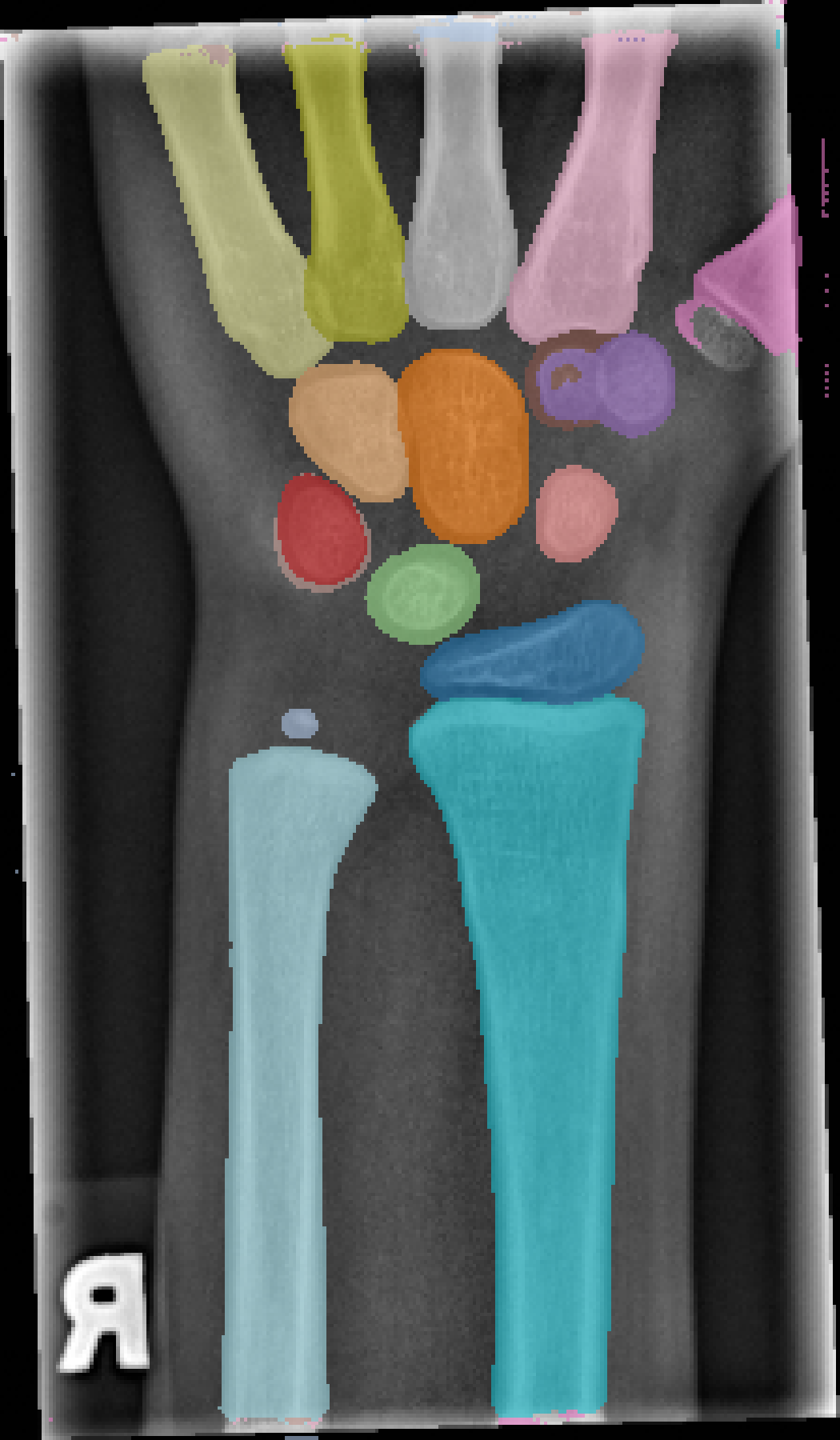}
        \caption{MT: $76\pm18\,\%$}
    \end{subfigure}\hfill
    \begin{subfigure}[t]{.23\textwidth}
        \includegraphics[width=\textwidth]{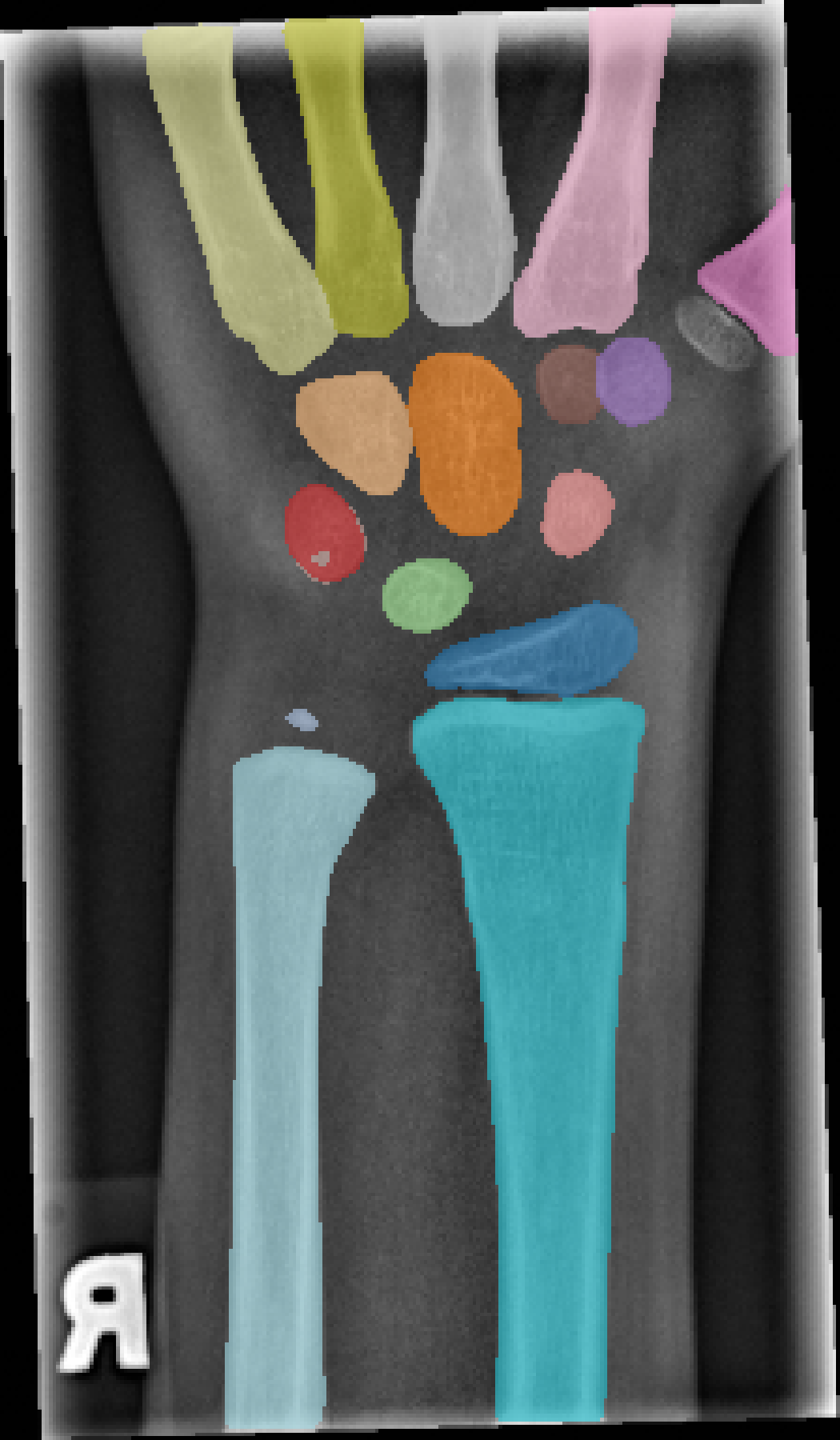}
        \caption{ours: $93\pm3\,\%$}
    \end{subfigure}
    \begin{subfigure}[t]{.23\textwidth}
        \includegraphics[width=\textwidth]{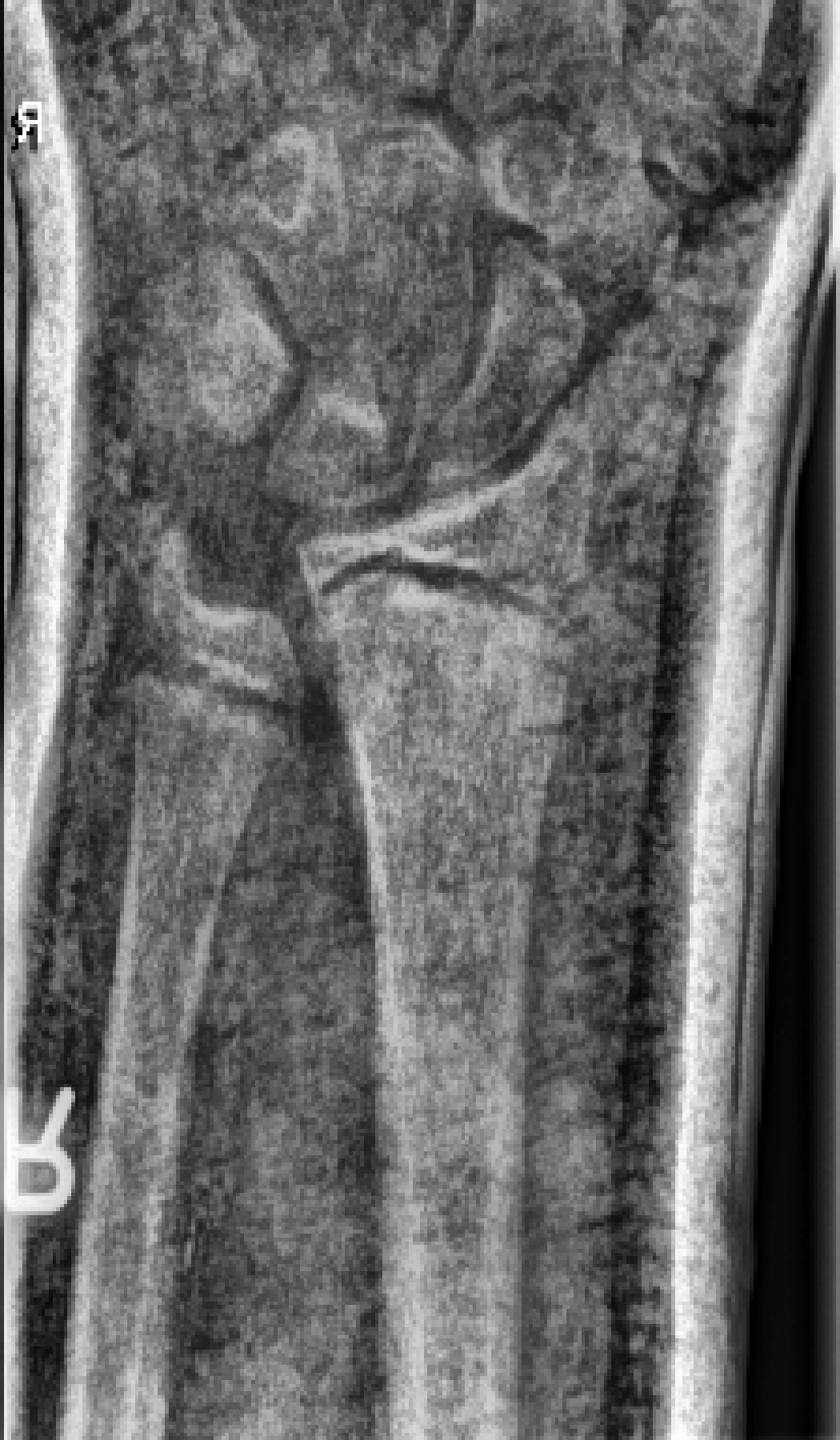}
        \caption{Worst test case}
    \end{subfigure}\hfill
    \begin{subfigure}[t]{.23\textwidth}
        \includegraphics[width=\textwidth]{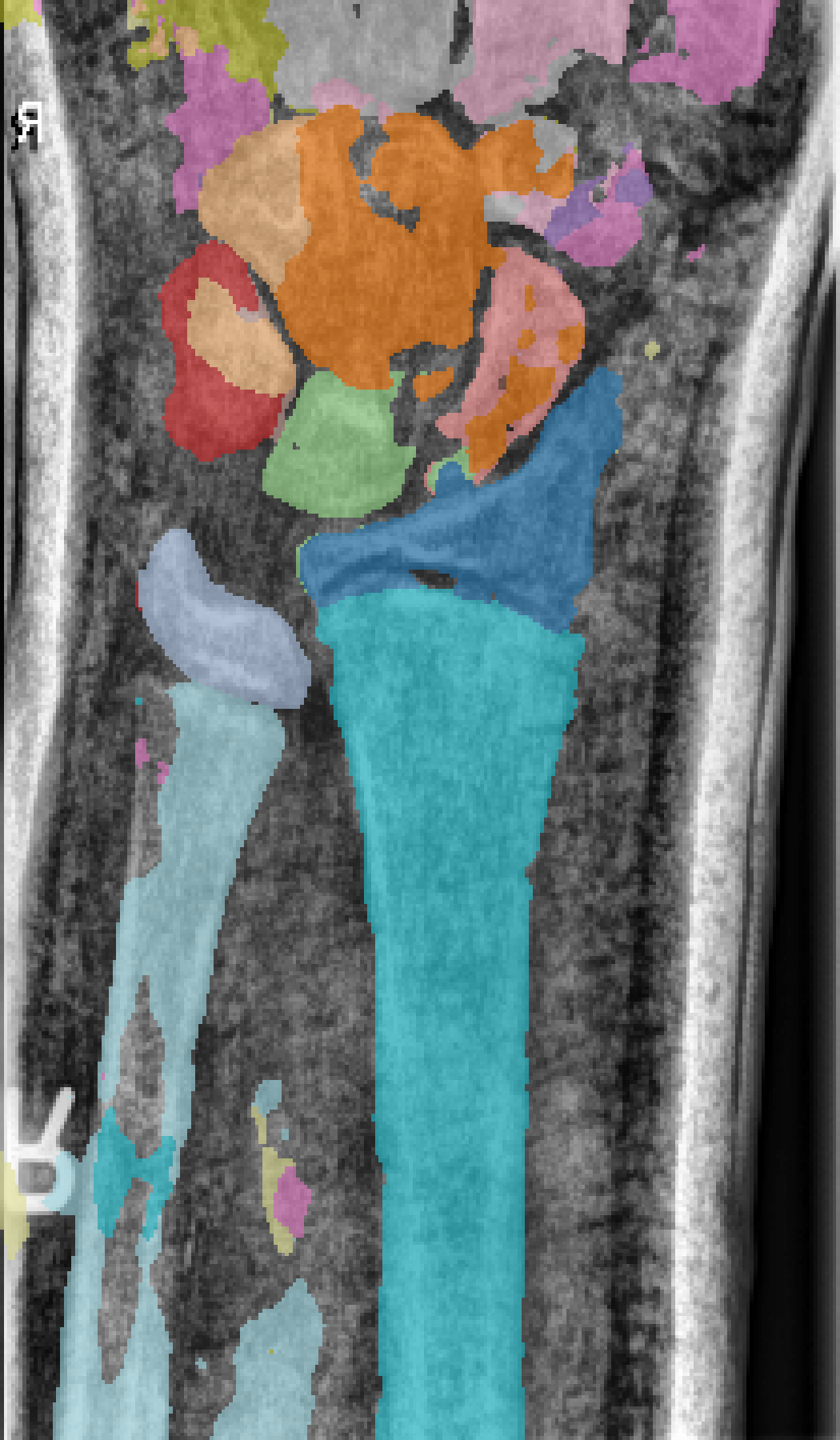}
        \caption{U-Net: $55\pm25\,\%$}
    \end{subfigure}\hfill
    \begin{subfigure}[t]{.23\textwidth}
        \includegraphics[width=\textwidth]{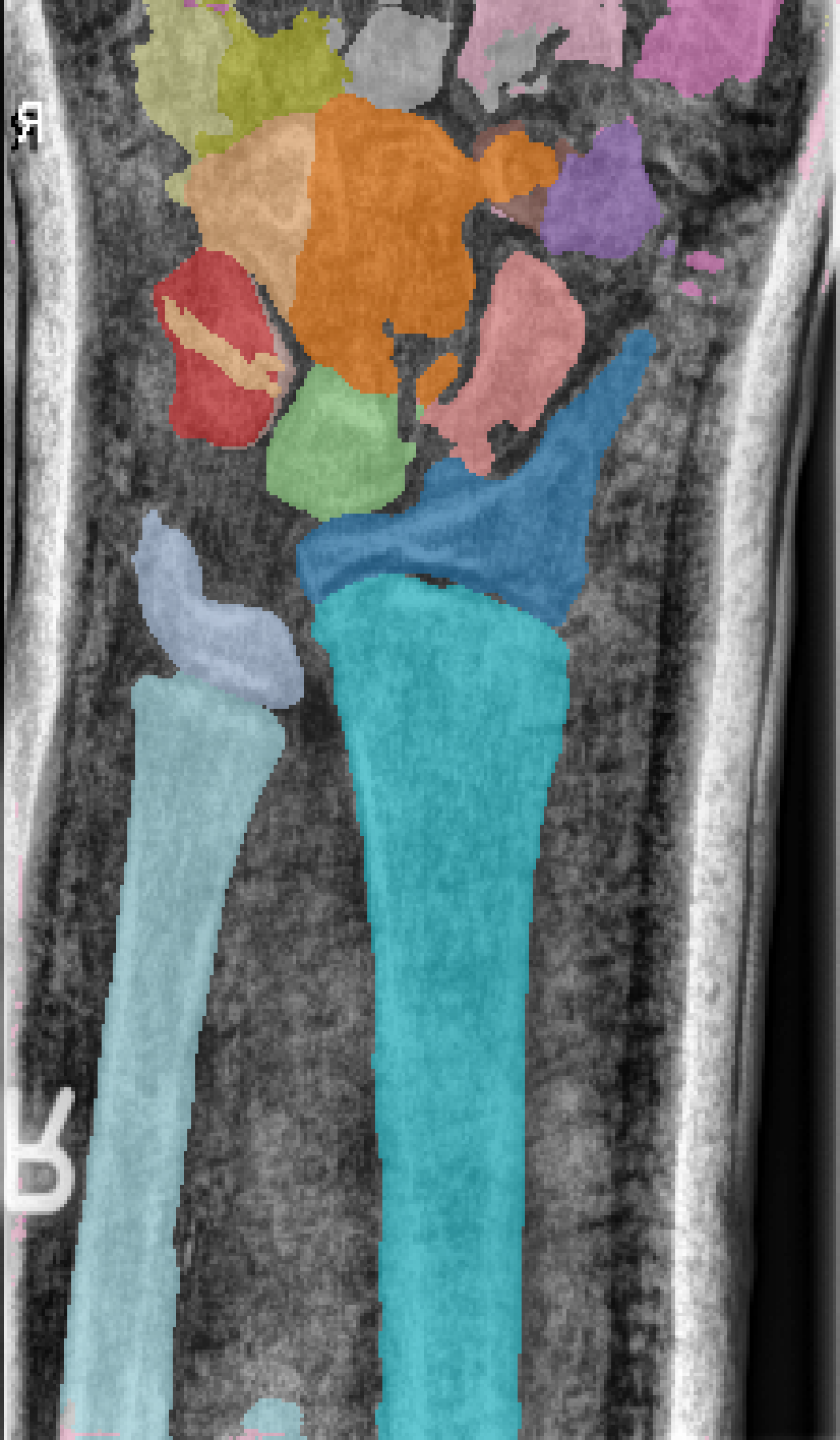}
        \caption{MT: $70\pm21\,\%$}
    \end{subfigure}\hfill
    \begin{subfigure}[t]{.23\textwidth}
        \includegraphics[width=\textwidth]{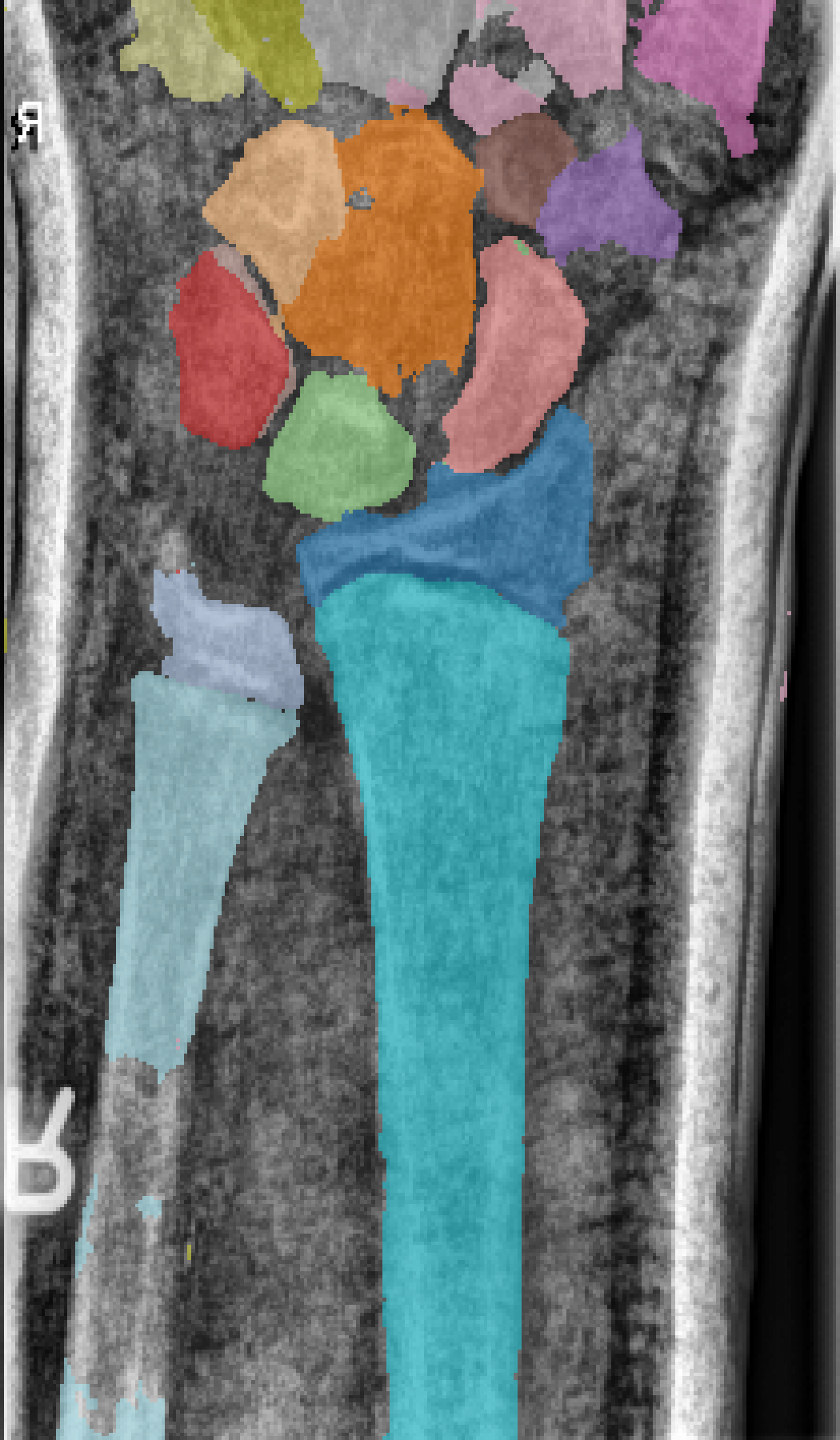}
        \caption{ours: $70\pm19\,\%$}
    \end{subfigure}
    \caption{First/second row shows the best/worst case of test split including DSC ($\mu\pm\sigma$) for a selection of methods. Second column: U-Net, third column: Mean Teacher (MT) and fourth column: U-Net trained with SAM refined pseudo labels on unlabelled data $\mathcal{Y}$ (ours).}
\end{figure}

\begin{figure}[h]
    \centering
    \begin{subfigure}{.47\textwidth}
        \includegraphics[width=\textwidth]{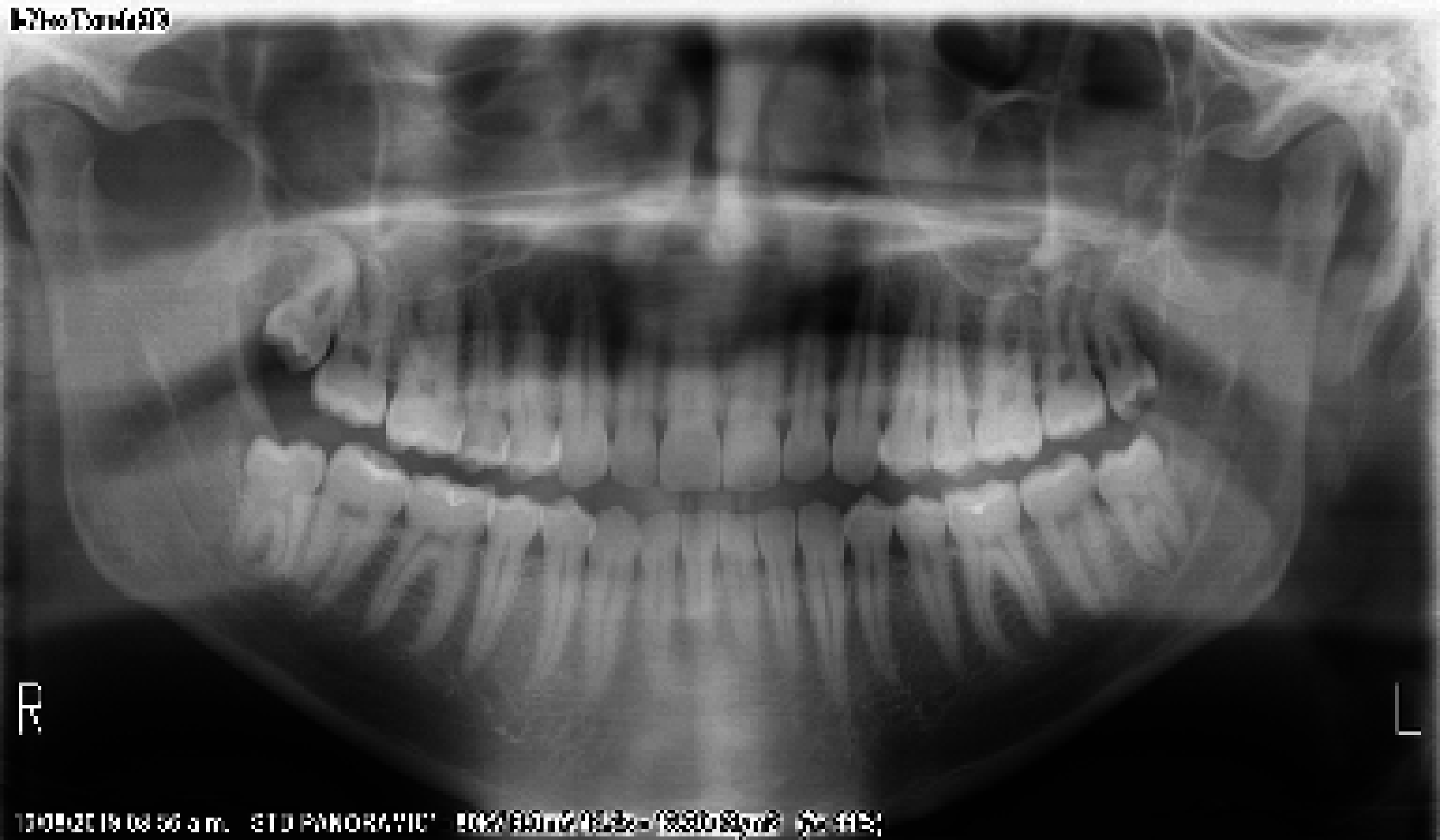}
        \caption{Best test case}
    \end{subfigure}\hfill
    \begin{subfigure}{.47\textwidth}
        \includegraphics[width=\textwidth]{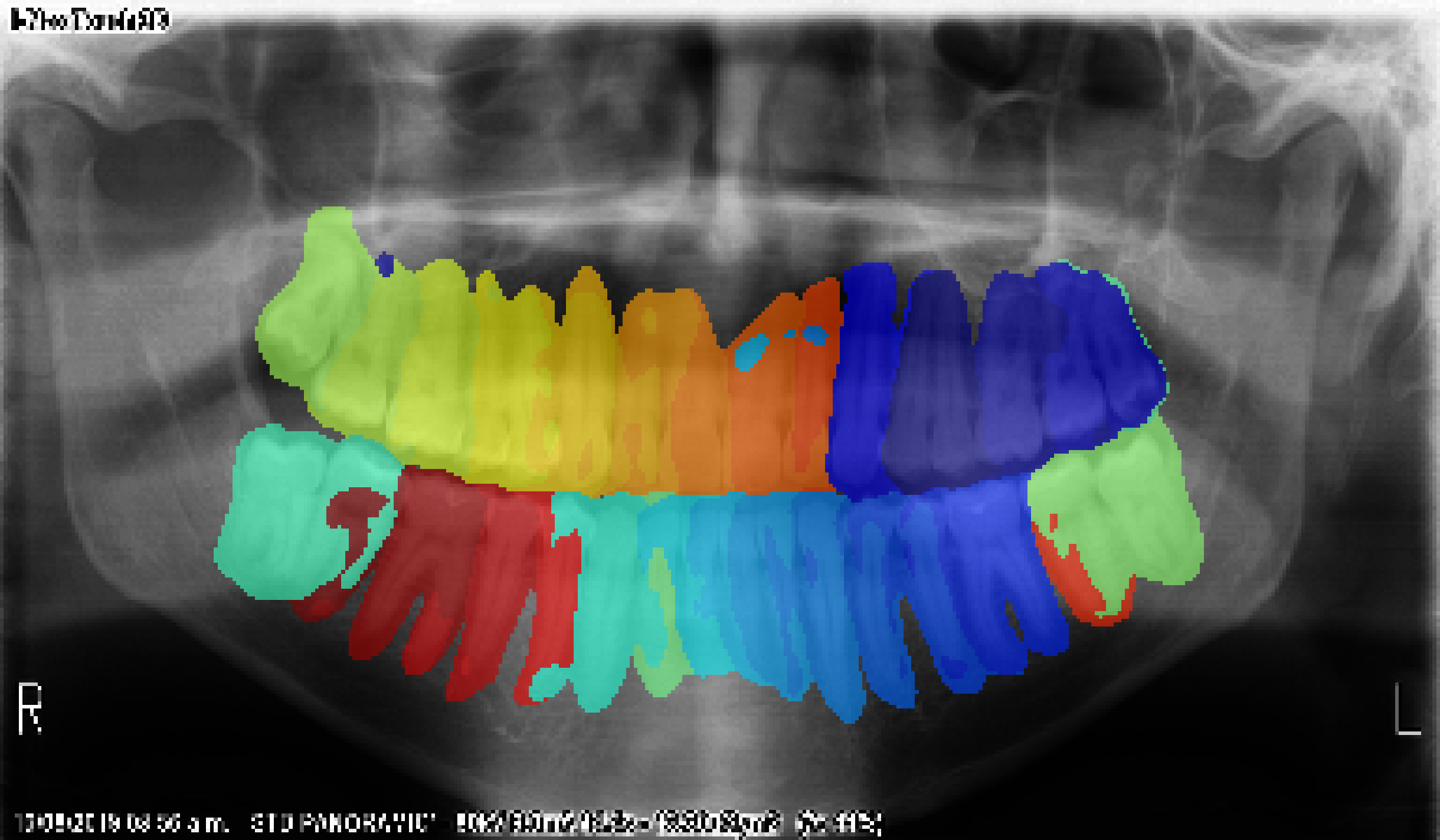}
        \caption{UNet: $61\pm10\,\%$}
    \end{subfigure}
    \begin{subfigure}{.47\textwidth}
        \includegraphics[width=\textwidth]{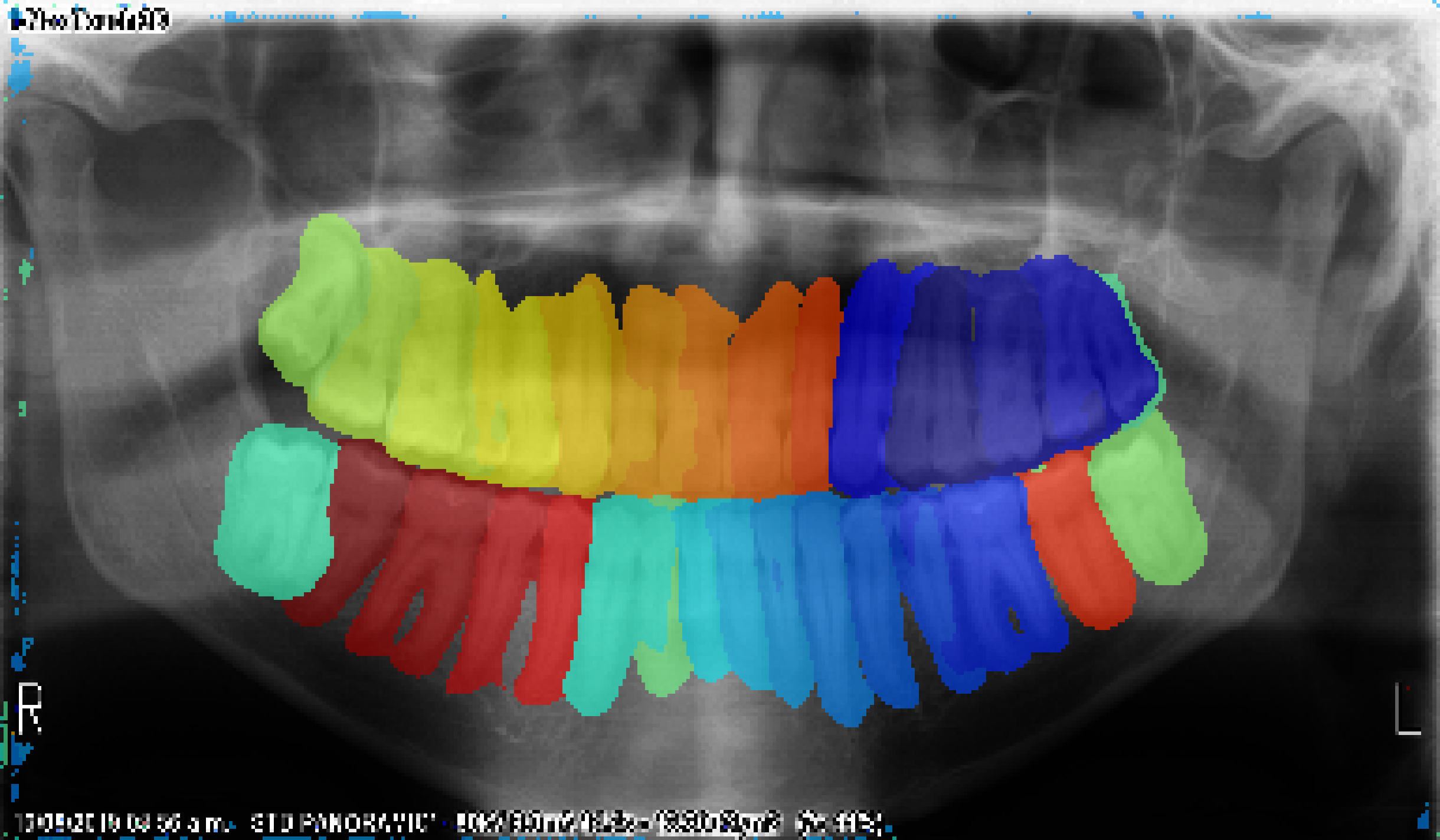}
        \caption{MT: $70\pm10\,\%$}
    \end{subfigure}\hfill
    \begin{subfigure}{.47\textwidth}
        \includegraphics[width=\textwidth]{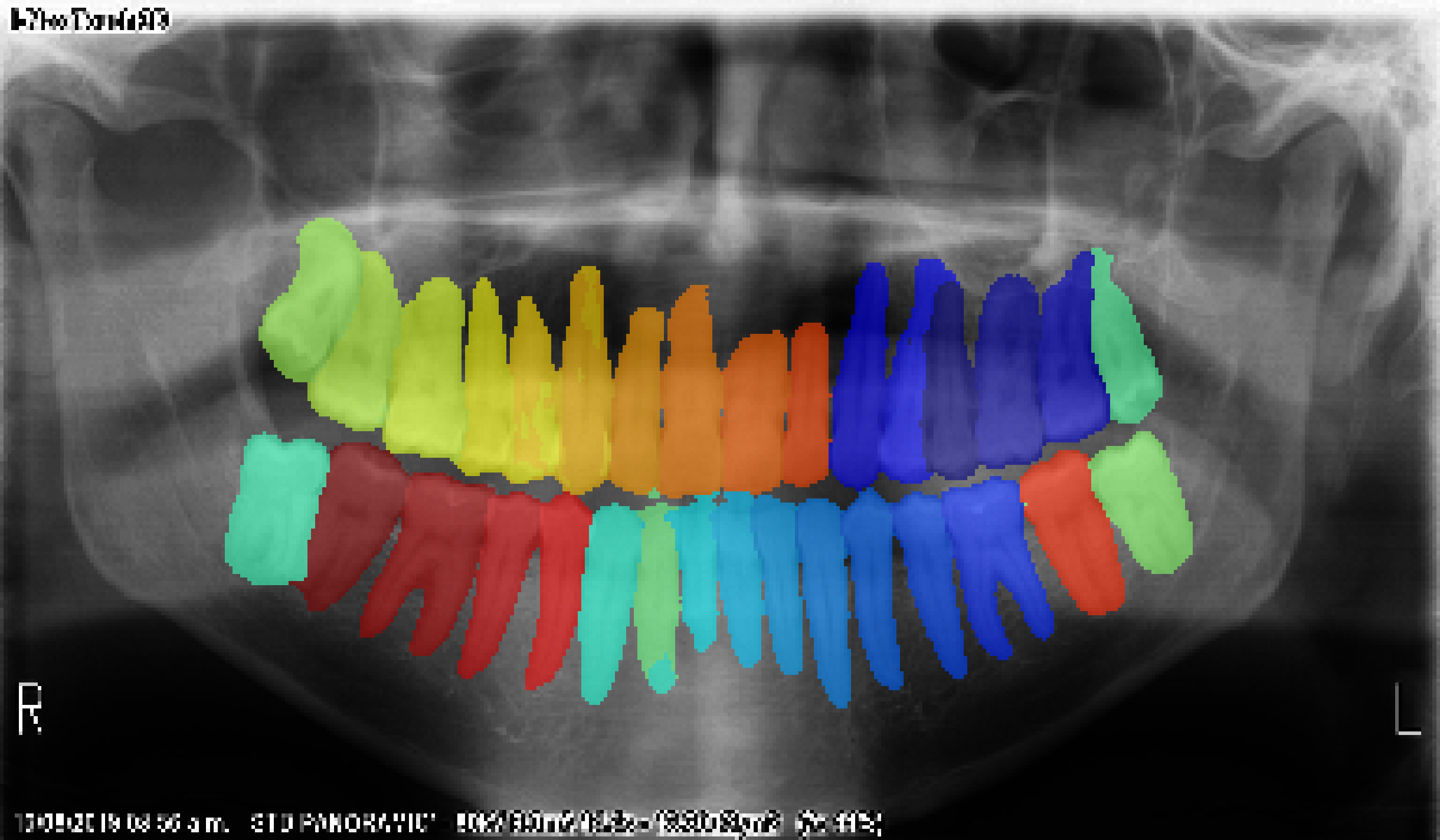}
        \caption{ours: $86\pm5\,\%$}
    \end{subfigure}
    \hrule\vspace{1em}
    
    \begin{subfigure}{.47\textwidth}
        \includegraphics[width=\textwidth]{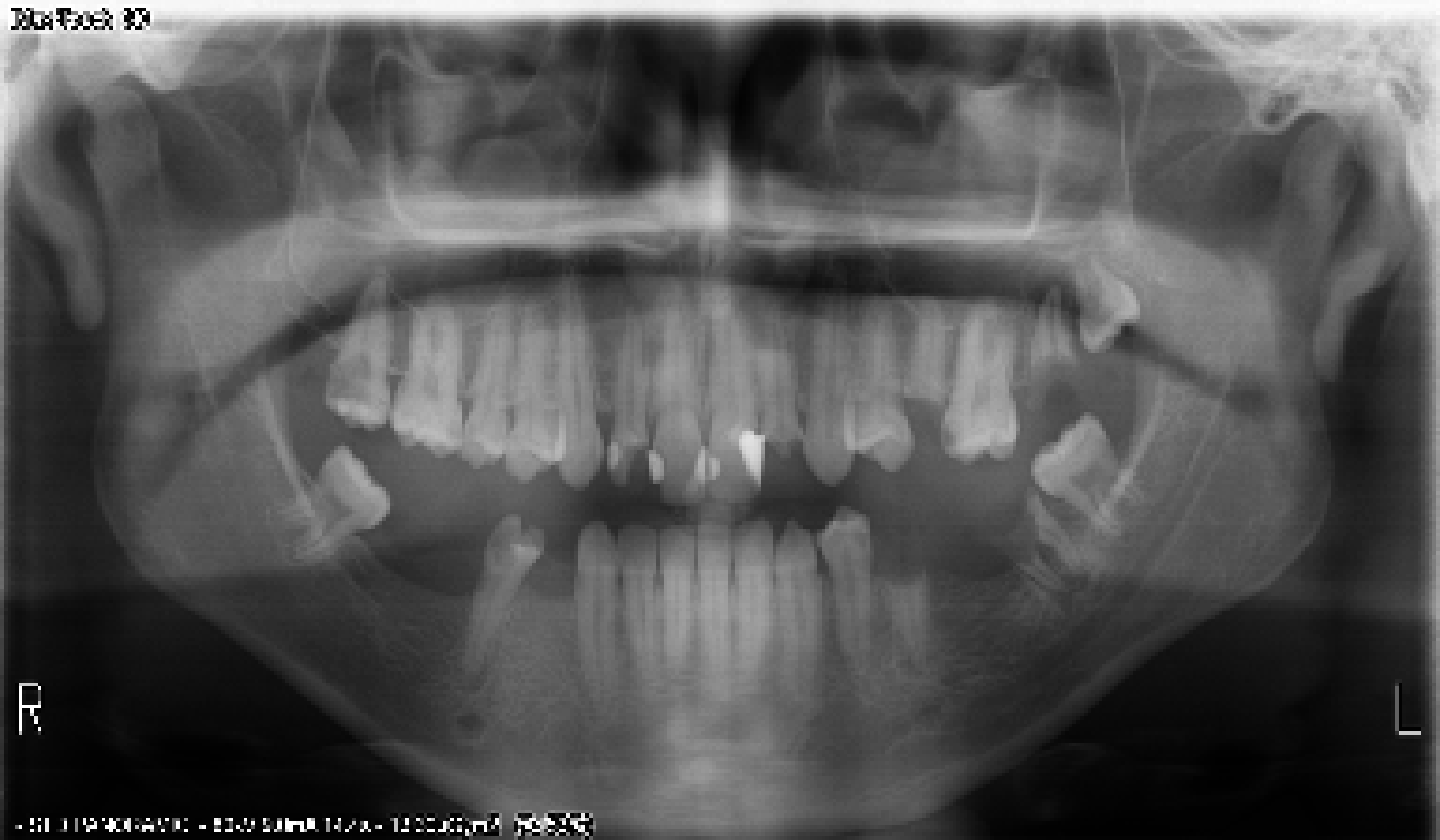}
        \caption{Median test case}
    \end{subfigure}\hfill
    \begin{subfigure}{.47\textwidth}
        \includegraphics[width=\textwidth]{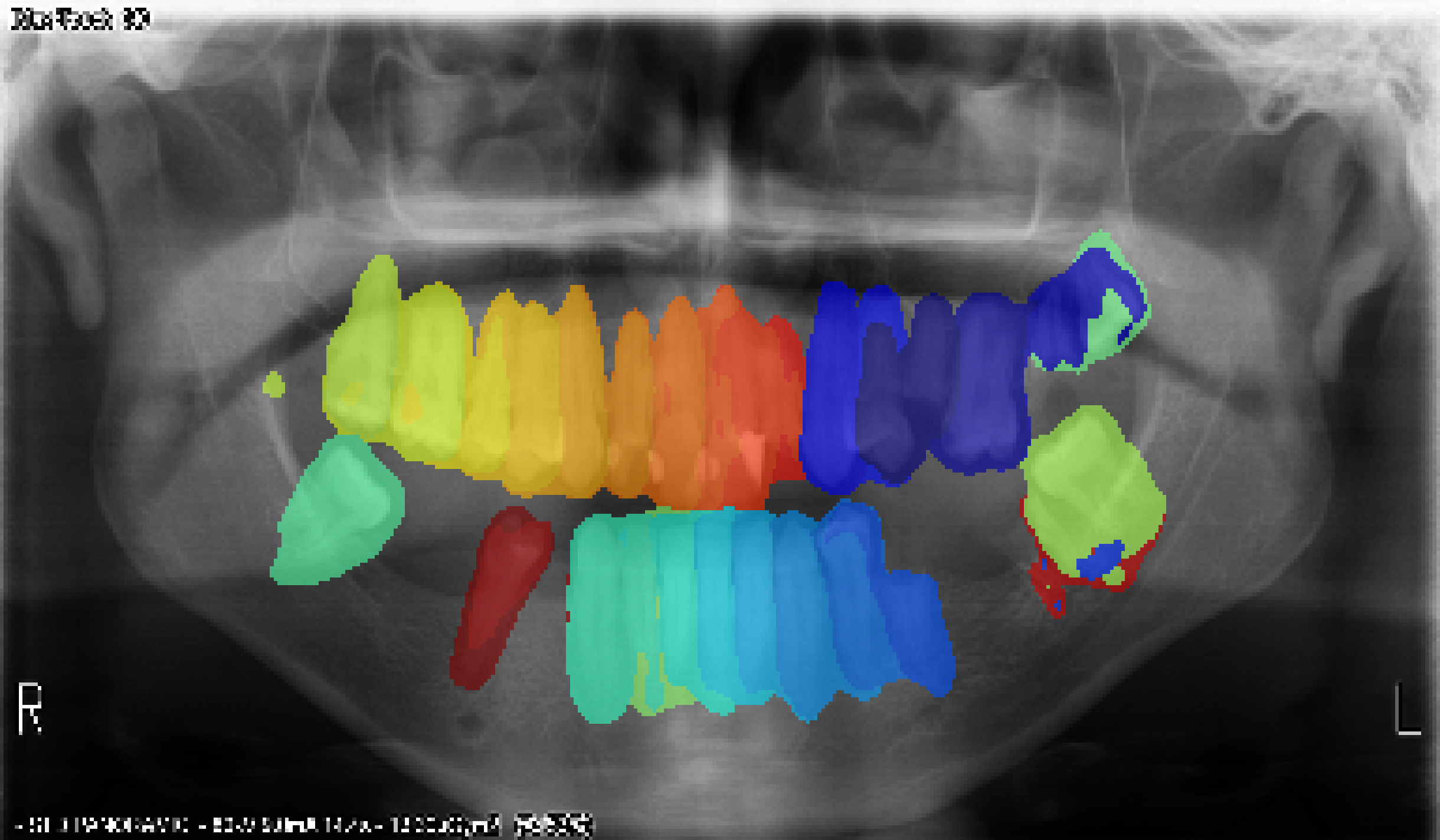}
        \caption{U-Net: $60\pm14\,\%$}
    \end{subfigure}
    \begin{subfigure}{.47\textwidth}
        \includegraphics[width=\textwidth]{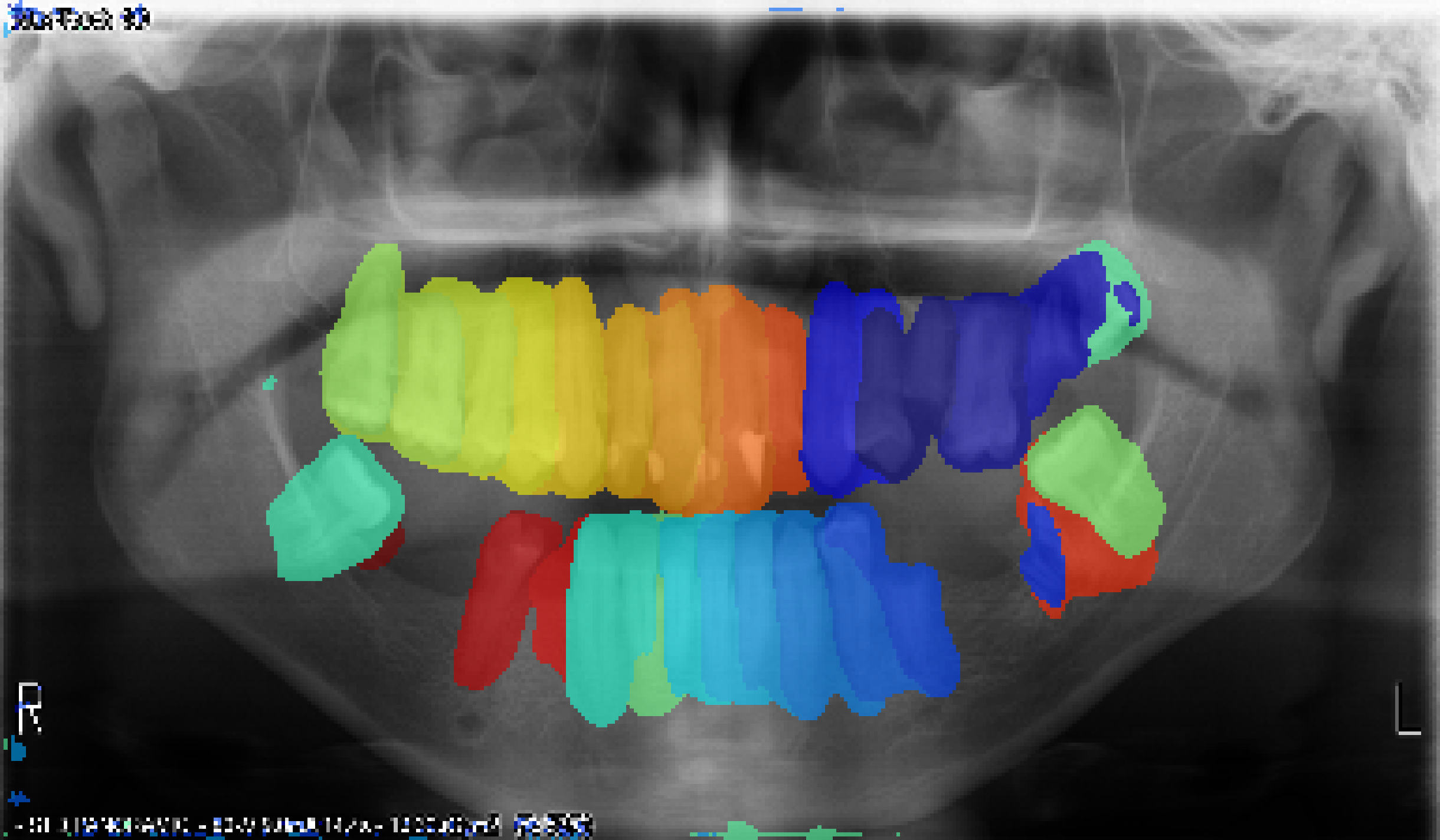}
        \caption{MT: $62\pm14\,\%$}
    \end{subfigure}\hfill
    \begin{subfigure}{.47\textwidth}
        \includegraphics[width=\textwidth]{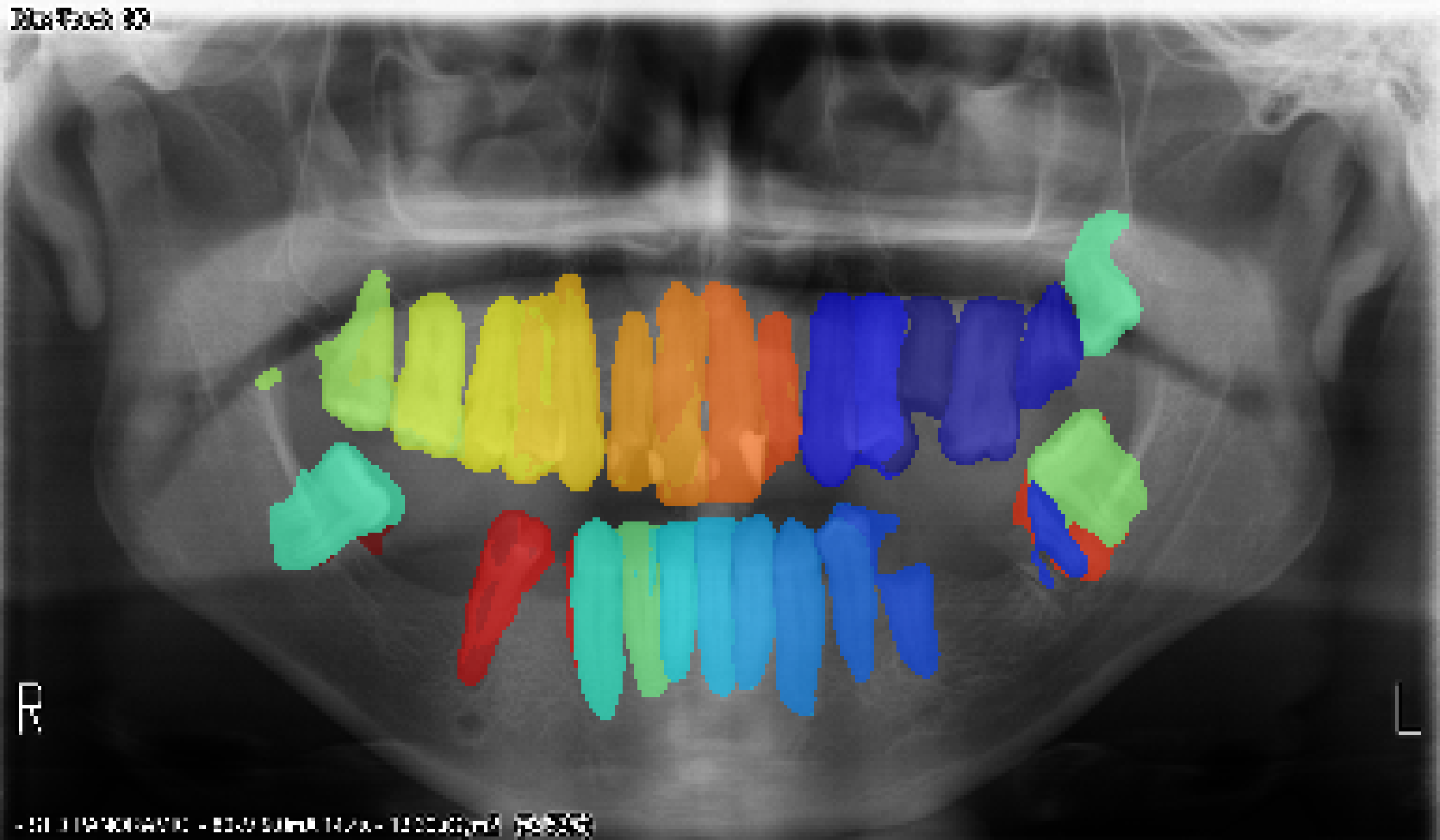}
        \caption{ours: $77\pm15\,\%$}
    \end{subfigure}
    
    \caption{First/second group shows the best/median result of test split including DSC ($\mu\pm\sigma$) for a selection of methods. "MT" describes the Mean Teacher and "ours" the U-Net $f_\varphi$ trained with SAM refined pseudo labels $\mathcal{R}$ on unlabelled data $\mathcal{Y}$.}
\end{figure}

\end{document}